# Advancing Assistive Robotics: Multi-Modal Navigation and Biophysical Monitoring for Next-Generation Wheelchairs

Md. Anowar Hossain[@*1], Mohd. Ehsanul Hoque[]
[@]Department of Electronics and Telecommunication Engineering
[&]Department of Electrical and Electronic Engineering
[*]Chittagong University of Engineering and Technology
[#]International Islamic University Chittagong
Chittagong, Bangladesh
[1]anowarhossain.cuet@gmail.com, [2]mohd.ehsanulhoque23@gmail.com

**A B S T R A C T**

Assistive Electric-Powered Wheelchairs (EPW) have proved not to be an optional tool but essential measure for people with disabilities such as ALS (Amyotrophic Lateral Sclerosis), Post Stroke Hemiplegia or dementia with mobility issues. The proposed system is developed to mobilize patients with such impairments with a novel multi-modal control mechanism that prioritizes the patients' needs and offers the ability to switch modes with ease. The four complementary control modes- Joystick, speech, hand-gesture and electrooculography (EOG) coupled with a continuous vital sign monitoring system (Heart-rate variability, $SpO_2$, Temperature), allows patients to be independent and caregivers to provide surveillance and earliest scope of intervention. Two-point calibration against clinical references for the biophysical sensors were conducted, resulting in a root mean square error of *≤2* bpm (heart rate), *≤0.5*°C (skin temperature), and *≤1%* for SpO2. Twenty participants with mobility related issues executed 500 navigation commands indoors, achieving command accuracies of 99% (joystick), *97%±2%* (speech), and *95%±3%* (gesture), with a loop latency of *20±0.5*ms. Caregivers receive real-time alerts via an Android application after encrypted cloud upload. By fusing multimodal mobility with cloud-ready health monitoring and documenting energy and latency budgets, the prototype addresses key gaps in assistive robotics, advances toward ISO 7176-31 and IEC 80601-2-78 safety compliance, and lays the groundwork for adaptive machine-learning enhancements in future iterations.

1. **Introduction**

Internet of things (IoT) has transformed the conventional approaches of automation and oversight in the recent years, impacting healthcare *[1] [2]*[1, 2], education *[3]* [3], infrastructure *[4] [5]* [4, 5], agriculture *[6]* [6] etc., thereby improving convenience and quality of life. Connectivity across heterogenous devices, superior computing capability and multi-layer perception *[7]* [7] models help establish task specific networks that can address domain-spread issues. Low-powered and wireless network connectivity enabled microcontrollers like ESP32, STM32 or the Raspberry Pi boards are some of the most widely used modules in IoT systems with communication protocols options like Wi-Fi *[8]* [8], LoRa *[9]* [9], NB-IoT *[10]* [10] and Zigbee *[11]* [11], BLE *[12]* [12] and provides robustness for data processing and communication. Biophysical sensor integration with such systems allows telehealth monitoring *[13] [14]* [13, 14] & non-invasive disease detection *[15]* [15] to operate as on-demand mobile healthcare solution.

Age related disease and disabilities are on the rise as there is report of the number of population aged over 60 will reach 2.1 billion by 2050 *[16]* [16]. A large quota of this demographic is burdened with neurological, musculoskeletal or musculoskeletal disparities such as Hemiplegia *[17]* [17], Amyotrophic Lateral Sclerosis (ALS) *[18]* [18], and Tetraplegia *[19]* [19] costing them their independence and quality of life. This portion of the world population require assistive technology like Wheelchairs or exoskeletal frames to help them regain marginal independence, but access to such technology is scarce in low to middle income countries, about 3% in general compared to higher income country residents *[20]* [20].

Electric Powered Wheelchairs (EPW) have alleviated caregiver burden and provided autonomy since the 1950s *[21]* [21]. First EPW invented by George Klein and later mass produced by Everest and Jennings' had analog joystick as the only maneuver choice, resulting in faster battery depletion, speed limitation by use of series resistors and longer charging hour of up to approximately 10h. On-Board 8-bit microcontrollers integrated wheelchairs showed significant improvement in battery backup, cutting power dissipation by 15% *[22]* [22]. Shared-control systems pushed the human-machine interaction in a forward direction, of which NavChair *[23]* [23] and Vision Chair *[24]* [24] uses vision based odometry for motion and navigation. Continuation in shared taxonomy in control interfaces researches later introduced Voice/Speech based control *[25]* [25], gesture control , Eye-gaze/EOG based control *[26]* [26] and Brain-Computer interface (BCI) control method *[27]* [27], which shaped the entire autonomous assistive wheelchair research landscape *[28]* [28] and projecting a 7.8% CAGR in the next ten year time frame *[29]* [29].

In the study, calibration for all sensors ensured <2% rms error rate across $SpO_2$, temperature, Heart rate variability, when compared to analogue devices. Control modes maintained over a 95% accuracy during clinical environment trials. The four control modes offered- joystick, speech recognition, gesture and EOG-based motion provides broader choice range for patients with different disabilities. In summary, this work bridges the predominant gap in current assistive wheelchair domain in :

1. Multiple input channels that cater to the user's physiological ability.

2. Two-point sensor calibration for medical grade sensors with <2% RMSE for health vitals measured.

3. Ad-hoc Wi-Fi communication protocol, offering .004ms loop latency during edge-IoT transmission.

4. Fail-safe control mode arbitration with latched stop for danger evasion and 10 hours battery backup ensuring long-stretch use case.

Xu et al. [30] *[30]* presented deployed a monocular-camera CNN with CBAM attention to decode octa-directional commands, involving a NVIDIA Jetson Nano for 1 ms travel. A joystick with trajectory-to-speed calibration adds indoor corner evasion capacity. However, the single modal control makes it harder for people with muscle fatigue.

Meligy et al.[31] *[31]* combines an SSVEP-EEG BCI interface with a joystick in a tri-wheel stair-climbing chassis. The on-boarded RHM setup streams HR, SpO2 and ECG to cloud and delivers alerts on anomaly detection. However, their system lacks a sensor calibration method and command latency in higher than acceptable.

Hussain et al. presents a EEG record while achieving 74% mental-state classification with an OvR logistic regressor *[32]* [32]. A wireless headset streams data to a PC that relays commands and a contingency switch hands control back to the joystick when stress spikes. It however shows limitation in signaling any vital changes in patients' body.

Ramaraj et al. *[33]* [33] bolt a Jetson-Nano stereo-vision pod and ROS move base onto a stock EPW, run RTAB-Map SLAM for visual odometry, and demonstrate collision-free navigation in Gazebo and hospital corridors . The system's modular rails avoid drilling the chair frame, and a recovery-behavior state machine unsticks tight hallway jams . Gazebo logs show 0.5 m s$^{-1}$ cruise through 80 cm gaps. *Limitation* – user input remains a single joystick; no vitals, cyber-security, battery or ISO compliance metrics are provided.

Hou et al.[34] *[34]* retrofit a commercial PC-201 chair with MySignals BP/HR/SpO$_2$/temperature modules, an Arduino data hub, and an ESP8266 that compresses and uploads vitals to ThingSpeak every 40 s while an ultrasound-CAN unit toggles between joystick and obstacle-avoidance drive modes. Yet the study omits end-to-end command latency, battery-drain profiling, and calibrated sensor error bounds, and its ultrasound array leaves "corner" blind spots that still trigger collisions—leaving both clinical validity and safety robustness unverified.

This work presents a complete solution aimed towards solving two major issues still persistent in the assistive healthcare technology, shared control interface for HMI operation and remote health monitoring (RHM) fusion. The work is an improved version of our project "Design and Implementation of an Autonomous Wheelchair *[35]* [35]", published in 2019. The previous work was focused on the control modalities only but the improved project deals with the edge connected Remote Health monitoring capabilities on top of the detailed and refined control mechanisms.

This paper contributes to limiting the gap in current landscape by fusing four control modalities- joystick, speech, hand-gestures and filtered EOG through an arbitration system with mode switching. On-board MAX30100 and DS18B20 sensors keep the HR and SpO2 error <2% and temperature error <5 °C. Dashboard capability and automated alert mechanism through email and SMS on Wi-Fi or GSM network provides quick intervention and round the clock oversight.

## 2. Methodology

The development of the assistive robotic wheelchair centers on integrating advanced technologies to simultaneously address mobility and health monitoring needs. The system design leverages Internet of Things (IoT) capabilities, a robust microcontroller architecture, and a suite of sensors to ensure adaptability, efficiency, and real-time functionality.

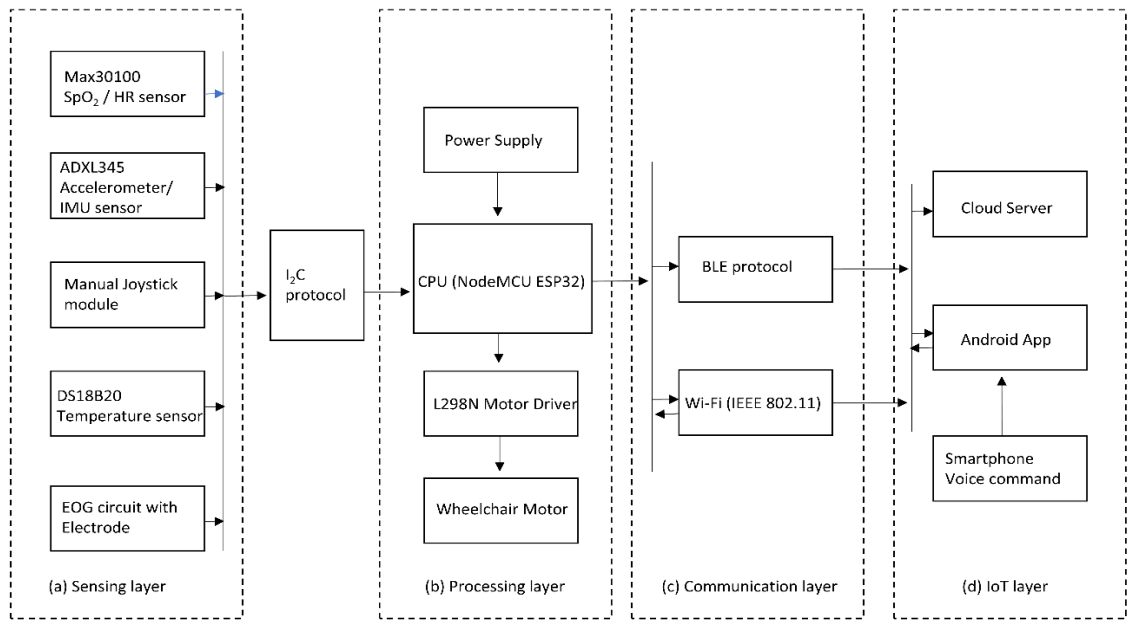

Fig. 1. Block Diagram of the System

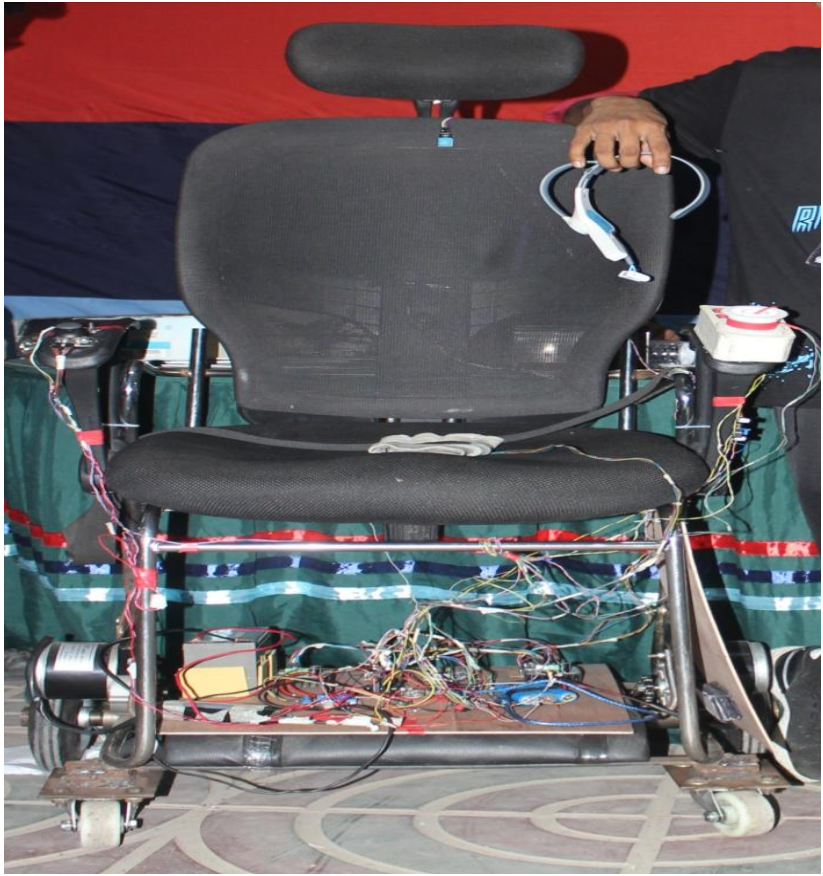

Fig 2: Real-time prototype with custom mechanical structure

## 2.1 System Overview

At its core, the system employs the NodeMCU micro-controller, which acts as the central processing unit (CPU). This module is responsible for interfacing with various sensors and control systems, executing decision-making algorithms, and managing data flow. The system combines mobility assistance with continuous physiological monitoring to deliver a holistic solution for individuals with diverse needs. A modular approach ensures flexibility, allowing the user to select specific control mechanisms and providing the caregiver with real time health insights. Data collected from the wheelchair's sensors are processed locally by the microcontroller and transmitted wireless to a cloud server for further analysis and remote monitoring. Figure 1 shows a complete system diagram. The wheelchair is equipped with a comprehensive health monitoring system that tracks key physiological parameters, including heart rate, oxygen saturation, and body temperature. Environmental and activity-based parameters, such as falls and convulsions, are also monitored through specialized sensors. These measurements are continuously logged, analyzed, and visualized through an Android application, enabling caregivers to respond promptly to critical events. Integration of IoT technology enhances the system's functionality by enabling seamless data transmission and remote accessibility. A cloud-based server aggregates sensor data and generates notifications in case of anomalies, ensuring timely intervention. The Android application serves as the primary interface for caregivers, providing real-time updates, alerts, and historical data trends.

## 2.2 Sensor selection and calibration

For the continuous health monitoring capabilities, a suite of biophysical sensors with advanced calibration mechanics and high precision were sorted out. The active sensors and microprocessor module provided real-time health monitoring with a minimal loop latency of 0.5 ms. MAX30100 is an optical module with PPG waveform emission and detection technology, offering clinical grade precision in the detection of blood oxygen saturation and heart rate, DS18B20 for skin temperature measurement, ADXL345 as IMU sensor working as gesture detection for wheelchair motion.

Table 1 justifies sensor selection while mentioning key features and advantages for clinical usage over other consumer grade sensors. Ensuring efficiency and cost effectiveness is key in developing such systems. Two-point calibration is required to mitigate the factory offset and gain errors for sensors in order to ensure precision. Because a small error margin in pulse oxymeter or temperature measurement could push a patient's perceived state from 'normal' to 'critical'. The MAX30100 is calibrated using the ISO 80601-2-61 proto col using a clinical pulse oxymeter simulator. The DS18B20 temperature sensor's calibration was done by submersion in ice water and boiling point water and ADXL345 accelerometer sensor calibration required a +1g/ −1g orientation shift on a two-axis jig.

| No | Sensor module | Key accuracy spec | Selection criteria |
|---|---|---|---|
| 1 | MAX30100 pulse oximeter sensor | ± 2 bpm HR, ± 2 % SpO2 (18-bit ADC) | Dedicated LED driver & ADC yield 6 dB better SNR than MAX30105; complies with ISO 80601-2-61 front-end note |
| 2 | DS18B20 temperature sensor | ± 0.5 (9–12-bit) | Factory-trimmed; < 0.1 self-heating vs. thermistor + external ADC combo |

| 3 | ADXL345 MEMS accelerometer sensor | 13-bit, 4 mg/LSB | No gyro ⇒ 35 draw; ± 1 g orientation enables linear two-point fit; lower offset drift than MPU-6050 |

Table 1: Medical sensor features

## 2.3 Decision-Making Framework

The wheelchair's operations are guided by a decision-making framework that prioritizes safety, efficiency, and user adaptability. Inputs from the sensors are processed by the micro-controller to determine appropriate actions, such as triggering alerts for health irregularities or executing movement commands. Fail-safe mechanisms are incorporated to prevent conflicts between different control inputs and ensure smooth operation. A priority-based control mode is set in place to ensure motion commands do not overlap with each other. **Algorithm 1** demonstrates the priority-based logic for the command prioritization.

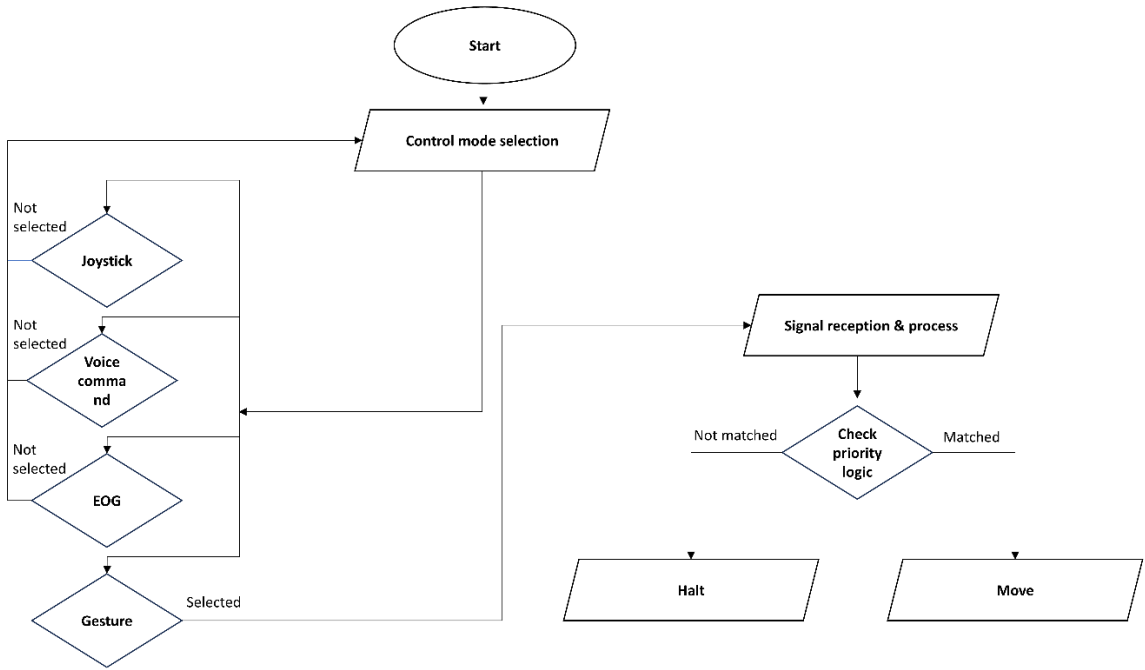

Fig. 3 Control mode selection flowchart

## 2.4 Data Flow and Alert Mechanisms

Data collected from sensors are transmitted via the NodeMCU's built-in Wi-Fi module to the cloud server. The server processes this information and makes it accessible through the Android application, where caregivers can view real-time updates and historical trends. Alerts for critical conditions such as abnormal heart rate, falls, or convulsions are automatically generated and delivered to caregivers in real time, enabling prompt action.

**2.5 Data transmission protocol**

BLE offers seamless short-range communication for smart medical devices using less power than typical communication proto cols available. For real-time monitoring, and data accuracy along with the most important aspect when designing an end-to-end data transfer model, robust security measure is ensured when using this protocol. Patient data such as identification, age or disease record are sensitive information only available to the healthcare providers and protected under certain legislative measures. IoT devices are prone to cyber-security threats and interceptions and countermeasures are necessary to protect such data. BLE offers AES-128/CCM encryption protocol while communicating between devices and the cloud, which provides the solution to block unauthorized access and protect user data. The ESP32 microcontroller used as the central processing unit offers built-in BLE capability, allowing RHM (Remote health monitoring) systems to run efficiently on-device. Additionally the low power consumption capabilities enable the encryption procedure to add 0.004 ms Round-trip time and less than <1% to the battery budget.

**Algorithm 1 Priority-ladder controller**

```
1: Input: JoySpeed, VoiceReady, GestureOK, EOGangle
FallFlag, HealthAlert, ObstacleFlag
2: Output: ActiveMode ∈ {Joystick, Voice, Gesture, EOG,
Stop}
3: SafeHalt ← false ▷ latched until hazards clear
4: while system powered on do
▷ — Hazard check —
5: if FallFlag ∨ HealthAlert ∨ ObstacleFlag then
6: ActiveMode ← Stop
7: SafeHalt ← true
8: else if SafeHalt = true then
9: ActiveMode ← Stop
▷ — Priority ladder —
```

```
10: else if |JoySpeed| > 50 and debounceOK(250 ms)
then
11: ActiveMode ← Joystick
12: else if VoiceReady then
13: ActiveMode ← Voice
14: else if GestureOK then
15: ActiveMode ← Gesture
16: else if EOGangle > 12° then
17: ActiveMode ← EOG
18: else
19: ActiveMode ← Stop
20: end if
21: executeMotion(ActiveMode)
22: wait 20 ms  ▷ 50 Hz loop
23: end while
```

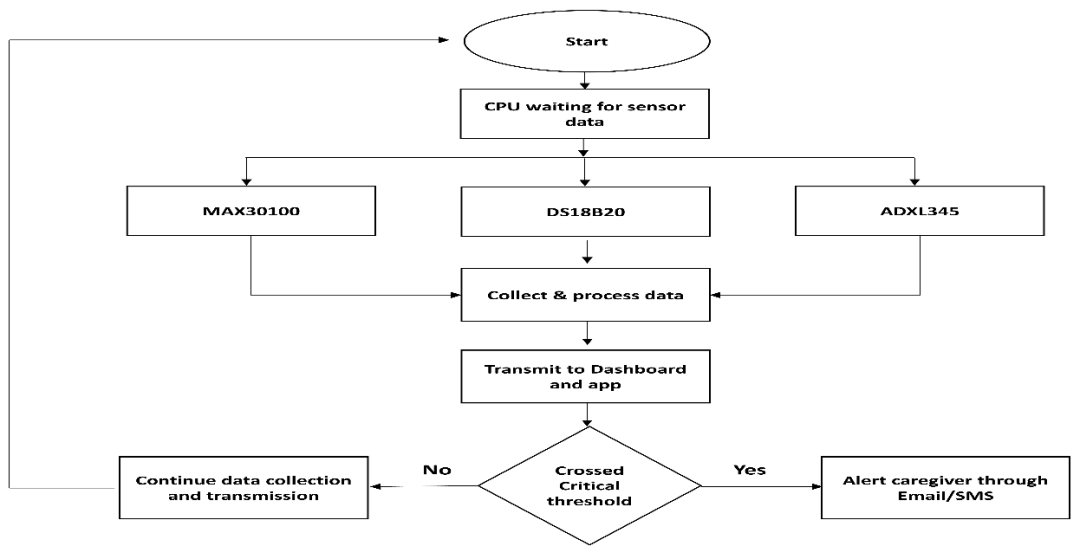

Fig. 4 Remote health monitoring flowchart

## 3. System Design and Control Modes arbitration

The system integrates mobility assistance with continuous health monitoring. A NodeMCU microcontroller acts as the central processing unit, coordinating inputs from multiple control systems and sensors, ensuring seamless functionality. The central operation unit is installed under the resting seat of the wheelchair.

### 3.1 Multi-Modal Control Deck

Key focus of this paper is the multi-modal control methods, which can be switched or cycled for the convenience of the user, with the help of 4 mode-switch buttons. The mechanisms are independent of each other, not causing any overlaps with an already operational mode. The four systems are the Joystick control, Voice operated mode, EOG (Electrooculogram) signal-based motion control and Gesture control mode using proper hand movements. The central processing unit has separate code-blocks to support each control method and a fail-safe mode to ensure interruption free command execution as demonstrated in Algorithm 1. FreeRTOS, a priority-based decision algorithm library custom made for the ESP32 microcontroller, allows simultaneous command reception and overlap mitigation. We also implemented a manual mode switching system using push buttons to alternate between the preferred mode of control, as shown in Figure 3.

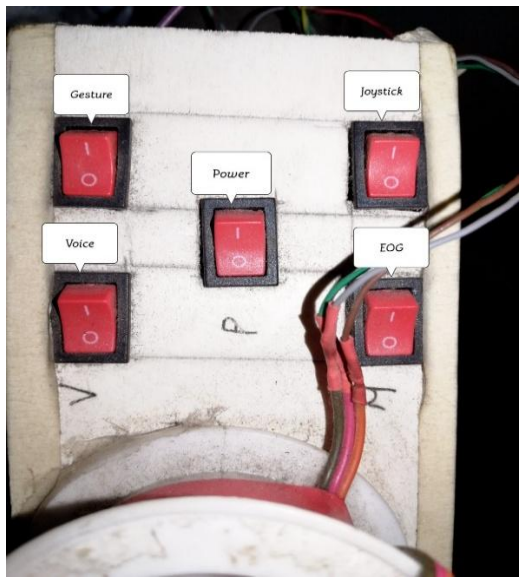

Fig. 5 Control mode selection buttons

### 3.1.1 Joystick Control

A manual control joystick is placed on top of the right arm chair for easy access. A joystick consists of several potentiometers distributed evenly across the X-axis and Y-axis, when the joystick is moved in a certain direction the potentiometers' resistance is changed which impacts voltage variation accordingly. The ana log signal is then transmitted to the ESP-32 microcontrollers input ADC (Analog to digital converter) pins. The two 18-bit ADCs' maps the analog input voltage ranging from 0-3.3V, to discrete integer values of 0-4095. The 12-bit resolution of the ADC means that the ESP32 can detect to digital values and further commands to move it in specific direction is executed. And a push button is stationed inside to sense when the joystick is being pushed, resulting in activation of the stationary mode of the chair.

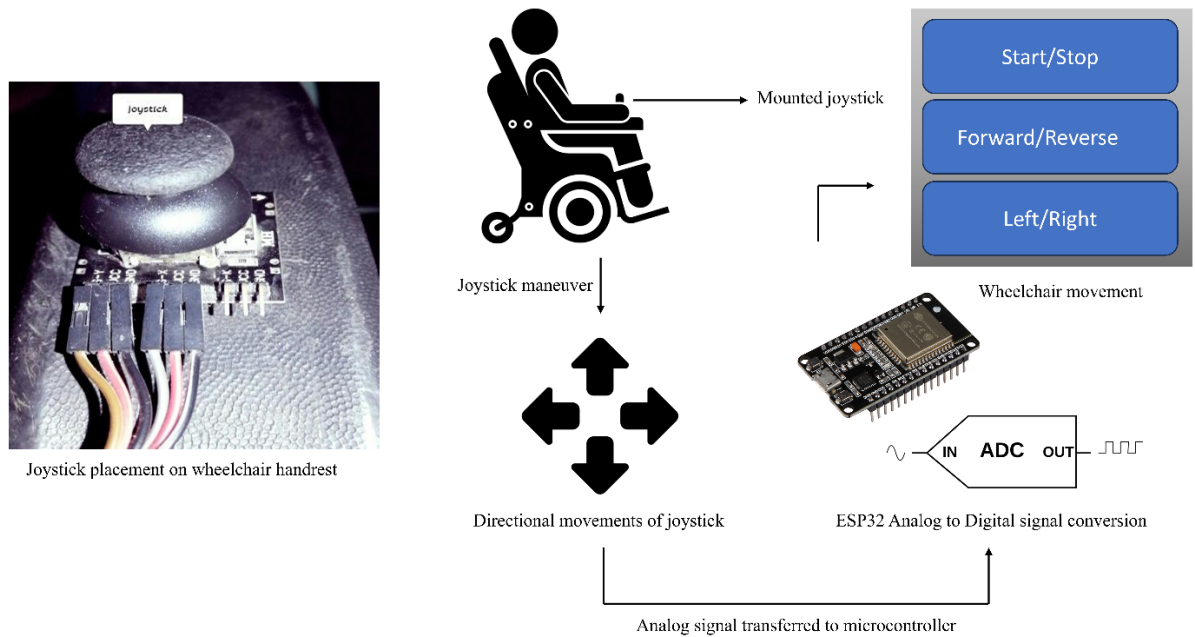

Fig. 6: Manual joystick mechanism

### 3.1.2 Voice Control

Voice operated locomotive devices uses short-range communication protocols to establish connection between devices and services. In this paper, the prototype uses a mobile application-based voice control system to accelerate and navigate the wheelchair. The workflow makes use of both the speech recognition service of a mobile app and the built-in wireless communication protocol, Bluetooth Low-Energy (BLE) of the ESP32. BLE was chosen for its low power consumption and reliable short-range communication, suitable for mobile wheelchair control. The application layer is the first in line to initiate the operation. The user speaks into the microphone of a smartphone and the voice input is received by the custom app. Android's speech recognition engine is then used to convert it into text, and specific grammar lexicons (e.g., "forward", "backward") are used to keep it from interrupting with other speech recognition services running in the smartphone, which are mapped to put the wheelchair in motion.

The text data is saved as string inputs and ready to be sent over to the ESP32. Android's Bluetooth-API is used to enable blue-tooth to scan for the ESP32 BLE host to establish connection. ESP32 uses a UUID(Universally Unique Identifier) to expose a ser vice and characteristic to send a command from the mobile app to the central processing unit. When a voice command is processed, the app will write the corresponding command data to the designated BLE characteristic. ESP32 acts as a blue-tooth host to listen to connections and data on the defined blue-tooth characteristics. A callback function is put in place to be triggered when data is received on the characteristic. The converted string input is received from the characteristic and the callback function decodes the command. Upon receiving and decoding the specific commands the ESP32 generates PWM and direction signals for the motor driver to move the motors to the user's desired direction.

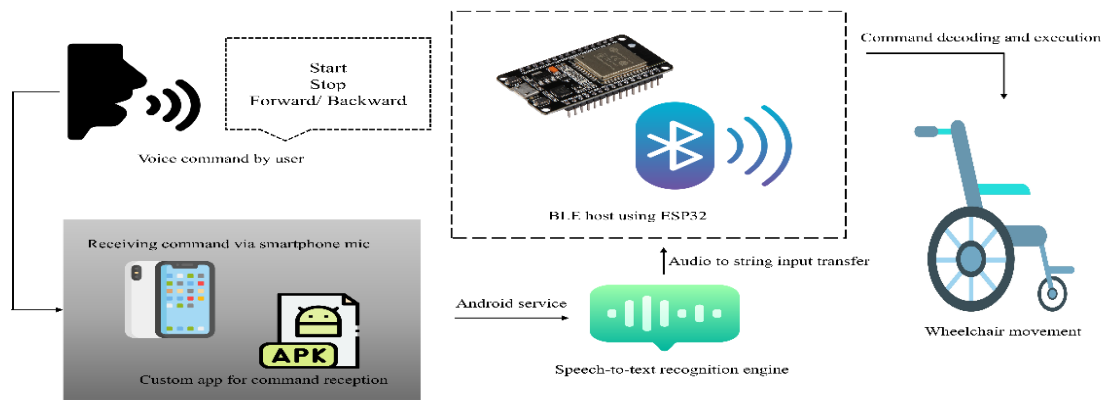

Fig. 7: Voice control mechanism using speech-to-text service

### 3.1.3 Gesture Control

Gesture control makes use of the ADXL335, a 3-axes accelerometer sensor used for tilting applications, as well as changes in dynamic acceleration due to motion, vibration, and shock. For people who face difficulty using a joystick or a voice-controlled app, gesture-based control provides an easier and more convenient solution. The sensor is attached to a hand glove for intuitiveness and hands free usability. The ADXL-335 consists of fixed plates as its base structure and moving plates with differential capacitors attached to them. Movement by the user results in output voltage variation proportional to the displacement. The output voltage is transmitted to the ADC pin of the ESP32 and converted to digital signal using its built-in ADC converter. Predefined gesture patterns (e.g., "Forward", "Backward") are mapped to the voltage variation that moves the wheelchair according to the user's hand gesture. For instance, if the user tilts his hands forward, the displacement of plates causes the capacitance to change eventually leading to the output voltage variation, and the signal is sent to the microcontroller to be acknowledged as command to move the wheelchair forward.

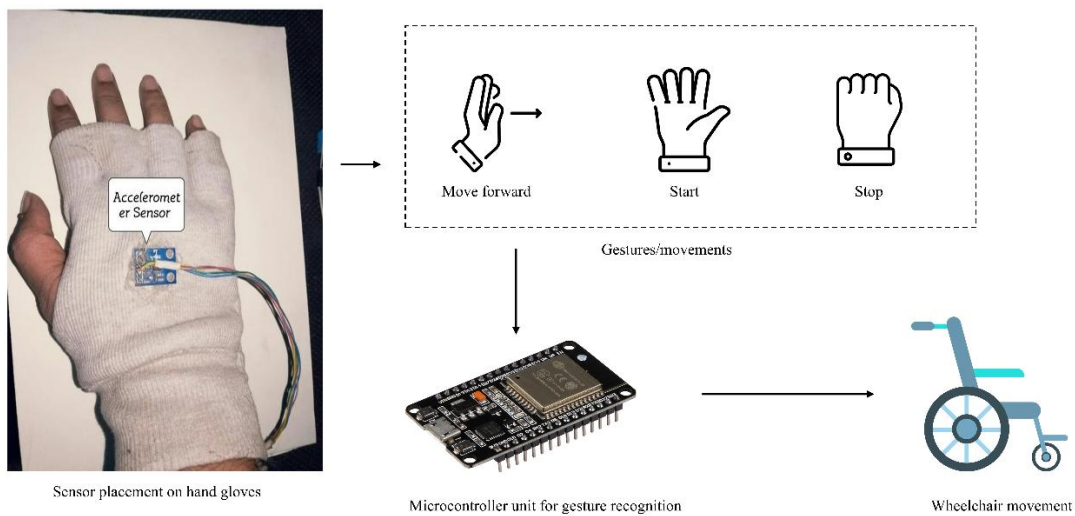

Fig. 8 Gesture control mechanism

### 3.1.4 Eye Control

The Electrooculogram (EOG) signal is another method used to control wheelchair movements for patients with quadriplegic dis order, where they lose control of all muscles except the ability to move their eyes and blink. For people with severe physical dis abilities, it is likely to find other control mechanisms difficult to maneuver and benefit from this mode of operation due to its limited control requirement and extensive usability. EOG measures the electrical potential difference arising from it being a dipole where the cornea (front of the eye) and the retina (back of the eye), where they are charged positively and negatively, respectively. The eye maintains a resting potential difference of 0.4 to 1.0 mV and varies depending on the change in dipole orientation whenever the eye moves horizontally or vertically. This change in potential difference is detected using the electrodes placed around the outer Canthi of both eyes for horizontal movement detection, and above and below one eye for vertical movement detection. EOG signals are very small, typically in the milli-volt range, that's why the signal is passed through LM358N, a dual-operational amplifier that amplifies the signal and filters out noise generated from other physiological activities like EMG, EEG etc. or from power-lines and/or electrode movement. This filtered signal is then communicated to the ESP32's ADC pin which con verts it into digital value and executes directional command according to the mapped values. The motor driver picks up the command and moves the WC accordingly. Each directional command is given a specific time interval to separate from the other ones. For instance if the patients stare in the forward direction for 4 seconds, the wheelchair moves in that direction. Or double blink of eyelids commences the stoppage of the wheelchair.

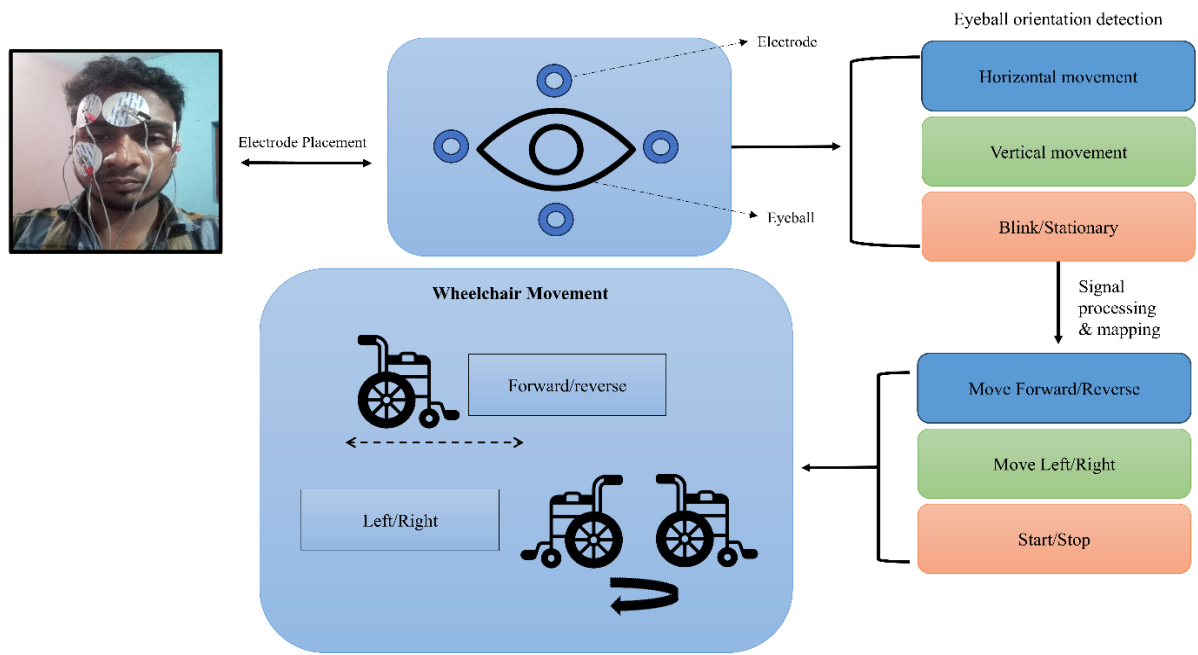

Fig. 9 Eye control mechanism using Electrooculogram

## 3.2 Patient Monitoring System

For continuous patient monitoring, this system incorporates a plethora of medical grade sensors. The patient monitoring system includes oxygen saturation, heart rate monitoring, body temperature, fall detection, heart attack detection, and convulsion detection. Several sensors have been used to monitor the patient. The sensors' information is sent to the server to store patient data for further monitoring. These sensors work independently with each other.

### 3.2.1 Oxygen Saturation, Heart Rate Monitoring and Heart Attack Detection

To monitor the oxygen saturation level in blood and heart rate the MAX30100 has been used. This module is an integrated pulse oximetry and heart rate monitoring sensor. It combines a photo detector, two LED's, optimized optics, and analog signal processing to detect heart rate and oxygen saturation. It operates in 1.8V to 3.3V power supplies. Blood oxygen saturation refers to the amount or volume of oxygenated hemoglobin the blood is transferring through the blood vessels. Both Oxy-hemoglobin and deoxy-hemoglobin has different light absorption properties and the pulse oximetry sensor makes use of that difference in absorption. The sensor module has a light source that emits infrared light and a photo-detector that senses the remainder lights after absorption. The transmitted light absorption is primarily dependent on the concentration of oxygenated hemoglobin and the light path or the width of the artery, complementing the Beer-Lambert law in physics. This module takes the signal from the fingertip and sends the data to the central processing unit. Then the signal is sent to the server and after that to the android application of the care giver. So, the caregiver can monitor the oxygen saturation level of blood and the heart rate of the patient. The sensor can also detect the heart attack of the patient. Whenever the level of the heartbeat of the patient increases greater than 140 or fall to less than 40, the micro-controller detects it as a heart attack.

### 3.2.2 Body Temperature Monitoring

Core body temperature of a healthy individual reflects their current metabolic state, accurate bodily functions and state of recovery if they are sick from any diseases. That's why it is essential to closely monitor how temperature variations occur in a person's body to account for the further actions in his treatment route. Generally, a healthy individual has a core body temperature within the range of 37 +/- 0.5°C (98.6 +/- 0.9°F), the temperature range needed for the body's metabolic processes to function correctly. For monitoring the body temperature of the patient, the DS18B20 has been used. This sensor provides an alarm function with user programmable upper and lower trigger points [DS18B20]. The temperature sensor has a resolution within the range of 9-12 bits and is usually powered in the 12-bits. With a low power requirement for operation, where it draws power from the data line, otherwise known as parasite mode, this sensor can sense temperature in the range of -55°C to +125°C. The A-D converter maps the temperature to digital values and the resultants are stored in the server and the mobile app for continuous monitoring.

### 3.2.3 Fall Detection

Falls can cause injuries, deaths, and hospitalizations of patients due to sudden injury. So, to detect the fall of a patient, we used ADXL-335. This accelerometer uses the x, y and z-axis values to detect any misalignment of the patient. Whenever a user trips or falls, the accelerometer sensor takes two types of measurement to detect a definite fall, they are respectively, the free fall due to gravity and impact with the ground. During free fall, the accelerometer senses an acceleration close to zero because of it being attached to the user and being relatively non-accelerating. The impact results in a significant spike in the voltage output as a result of the sudden and heavier orientation change in the 3-axis measurement capability. The near-zero acceleration period and sudden acceleration change are key indicators in detecting a sudden fall. Whenever sensor values cross their critical value, the microcontroller detects the condition as a fall of the patient and informs the caregiver through the android application.

### 3.2.4 Convulsion Detection

Convulsion is the rapid contraction and relaxation of the muscles due to illness such as high fever or severe underlying neurological conditions like epilepsy and seizure. Both have been termed a life-threatening neurological condition by the World Health Organization (WHO), that affect over 50 million people around the globe, causing them to be affected negatively in their social, educational, professional, and family lives. Earlier detection and intervention of these conditions can elevate the burden off the patients, ensuring a healthier and danger-free life. During such episodes, a patient will experience involuntary and rapid jerking movements in his muscles and other regions of their bodies, depending on the severity of the attack.

Vibration sensors are a type of accelerometer sensors that can detect sudden acceleration changes due to their piezoelectric properties. The mechanical stress induced by the patient is converted to electrical signals by the sensor. To accurately measure the convulsion/seizure episode, the vibration sensor is used in conjecture with other sensors arrangement, which are the Heart rate variability, SpO2 and accelerometer sensors. Vibration sensor is attached to parts of a patient's body where the abrupt shaky movements are most likely to be noticed, such as their hands or chest. The amplitude, frequency, and acceleration are measured key indicators to determining the severity of the seizure episode. Proper detection of such episodes involves measuring the acceleration changes to a certain threshold value predetermined by the detection algorithm, so the sensor senses the variation in value and sends a signal to the microcontroller unit. The ESP32 then sends an alert to the caregivers' phone through server and the monitoring app also receives an alert to check in on the patient at earliest interval.

### 3.3 Android Application

The healthcare app was designed using the Thunkable App inventor platform's unique and intuitive drag and drop customization options. The app consists of a dashboard and a command interface that, respectively, works on showing the health vitals of the patients and sending necessary commands to the wheelchair for moving. There is also a dedicated alert system panel where the caregiver can receive immediate notifications of any unexpected events, such as fall or seizure at any moment. Sudden spikes or irregularities in any or all health parameters are also notified through the app, and the caregiver can take the necessary measures to prevent fatal incidents. The ESP32 microcontroller unit receives the sensory data input and detects any anomalies through unique algorithms for each detection operation. Then the signal is sent to the app using Bluetooth Low Energy (BLE) to the smartphone app and the app takes the input to alert the caregiver by transmitting the condition and location of the patient.

### 4. Experimental Result and Analysis

The four control modes and health vitals' accuracy was measured and validated by employing 20 patients, age ranging from 50 through 80 years, with proper consent at a private owned hospital in Bangladesh. The patients tested the input modes and health sensor's accuracy under strict medical supervision in a clinical environment.

## 4.1 Gesture Accuracy

The autonomous wheelchair contains four control systems to increase the usefulness among disabled patients. Multiple supervised tests have been performed to evaluate the effectiveness of wheelchair control. The hand gesture controller has been tested for movements of the wheelchair toward forward, reverse, right, left, and halting. The voice control, eye control, and joystick control systems were also validated with the five commands. The voice control was performed at home environment system. The eye control system is most suitable for those who have no complexities in their eyes. Trial results were collected and shown in Table 1, where each control methods went through 100 trials each and based on that, the movement accuracy was logged and analyzed. The average movement accuracy across all control mechanisms were approximately 95%.

Table 3: The performance of four control modalities

| Command Name | Trial No. | Gesture Success | Gesture Acc (%) | Voice Success | Voice Acc (%) | Eye Success | Eye Acc (%) | Joystick Success | Joystick Acc (%) |
|---|---|---|---|---|---|---|---|---|---|
| Right | 100 | 95 | 95 | 90 | 90 | 95 | 95 | 100 | 100 |
| Left | 100 | 100 | 100 | 95 | 95 | 95 | 95 | 100 | 100 |
| Forward | 100 | 100 | 100 | 100 | 100 | 95 | 95 | 100 | 100 |
| Backward | 100 | 95 | 95 | 95 | 95 | 90 | 90 | 95 | 95 |
| Stop | 100 | 90 | 90 | 100 | 100 | 95 | 95 | 95 | 95 |

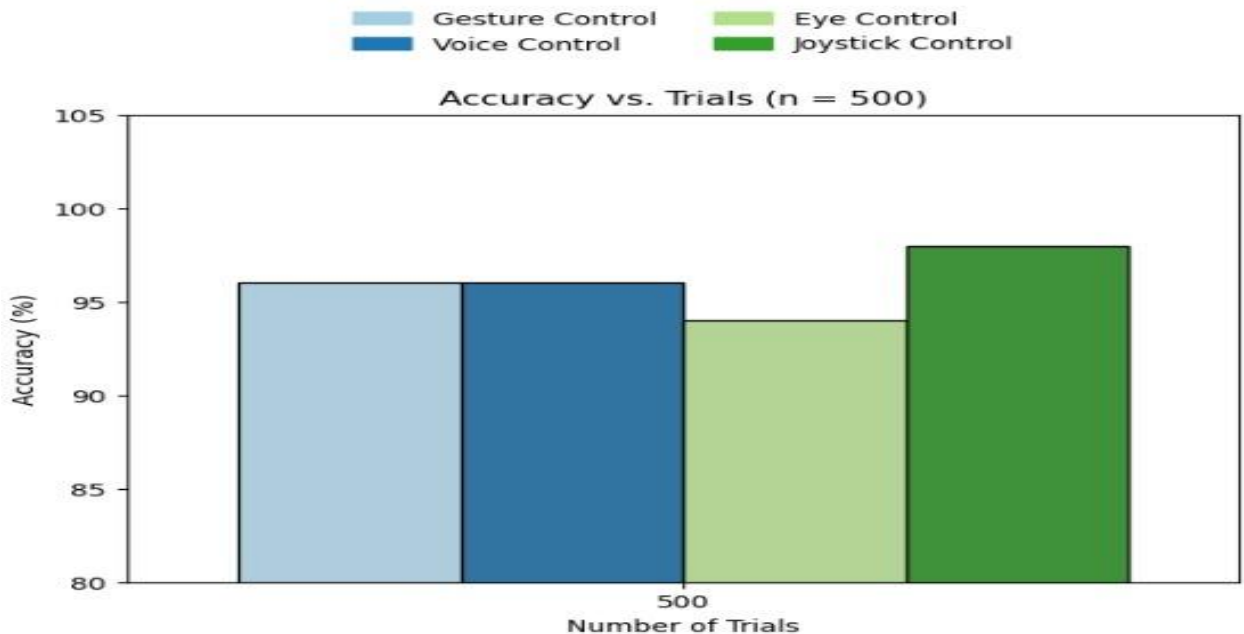

Fig. 10 Control mode accuracy bar chart

## 4.2 Server-side representation and alert system effectiveness

The system transfers all the data to the web-server named ThingSpeak which is an analytics platform for IoT. This allows to aggregate, analyze, and visualize the data. The Mobile app, enhanced with a compact and user-friendly interface shows the patients' health information, both the vitals' current conditions and alert systems. The vitals' show patients' body temperature, oxygen saturation and heart-rate variability in real time, while the alert systems are separated by a unique identifying colored button, Green and Red, to alert the caregivers of any unexpected situations like heart attack or convulsion etc. The Green alert typically indicates a normal or stable condition and red alert, after crossing or touching a certain threshold signifies the significant health severity. There is also the email alert, where a notification is sent to the authorized healthcare personnel in case of an emergency, with the date and time of the emergency via Gmail.

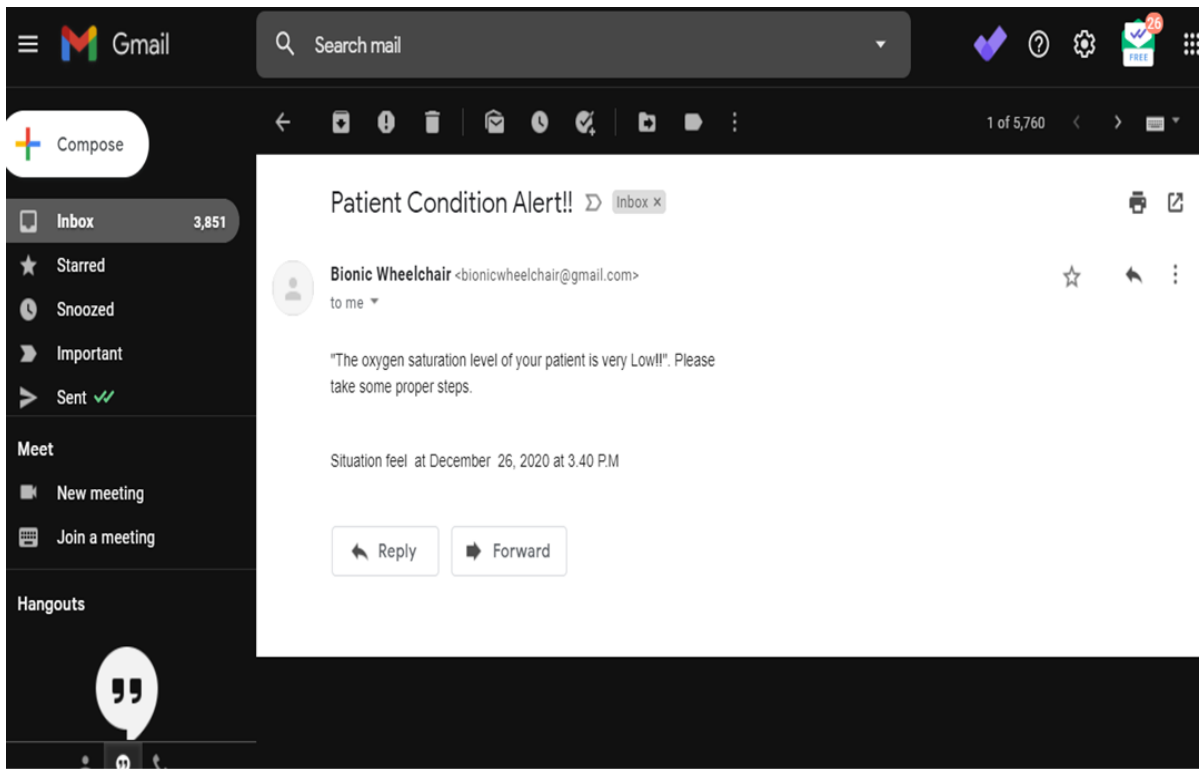

Fig. 11 Critical condition alert using email service

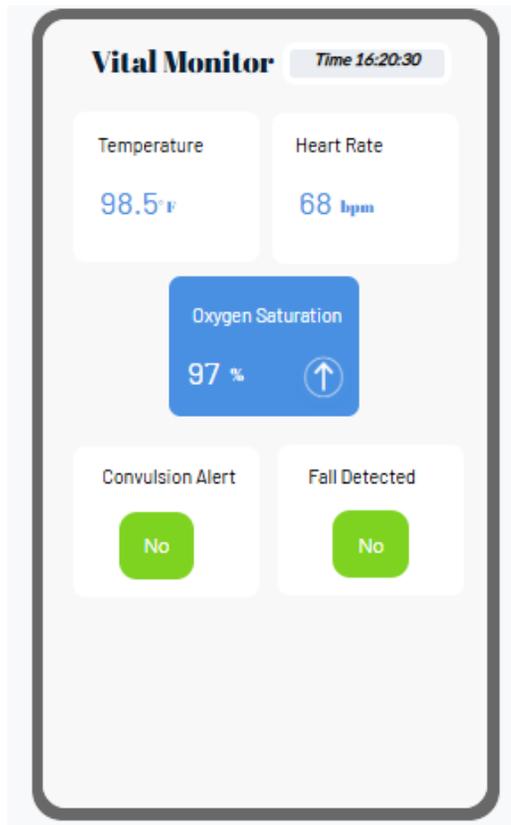

Fig. 12: Critical condition alert using mobile app interface

### 4.2.1 Heart-rate (BPM)

Figure 12 shows the BPM graph over the patients monitoring period in real time. Here in the server-side GUI, the continuous heart rate data is shown over the monitoring period, and logged in MySQL server for further analytics. A numerical figure is also shown for instant check-ins and the alert option is present for immediate notifying ability.

For BPM measurement, PROMIXCO patient monitoring device was chosen by the hospital authority for its medical usage certification, wider range of measurement and robustness. We measured our MAX30100's in-built HR-BPM sensor against the PROMIXCO monitoring device to determine its accuracy. We selected one patient for the trial.

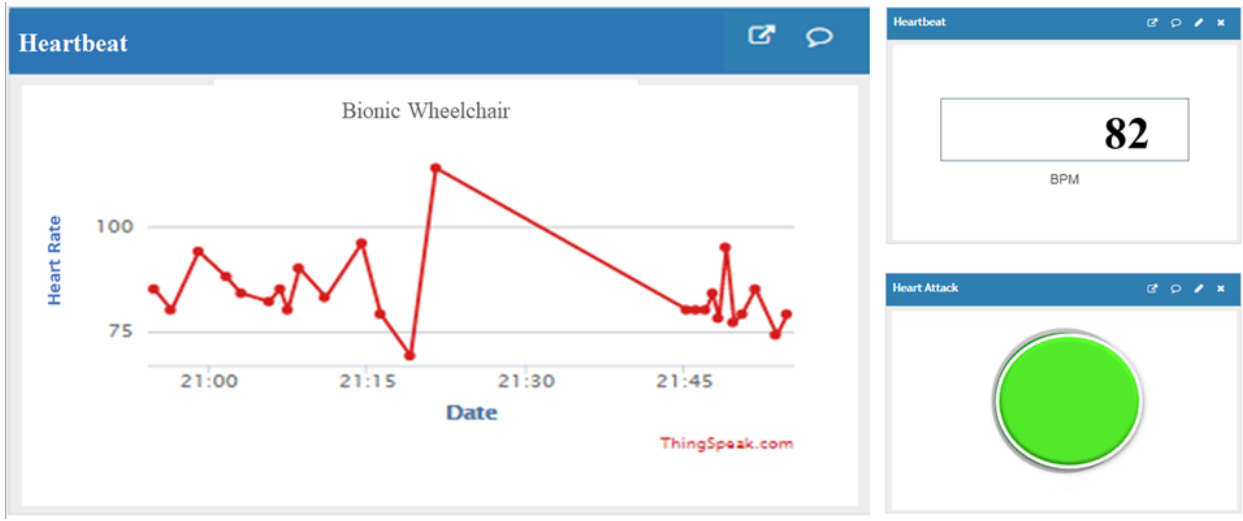

Fig. 13: Real time Heart-rate BPM tracking and alert monitor on ThingSpeak server

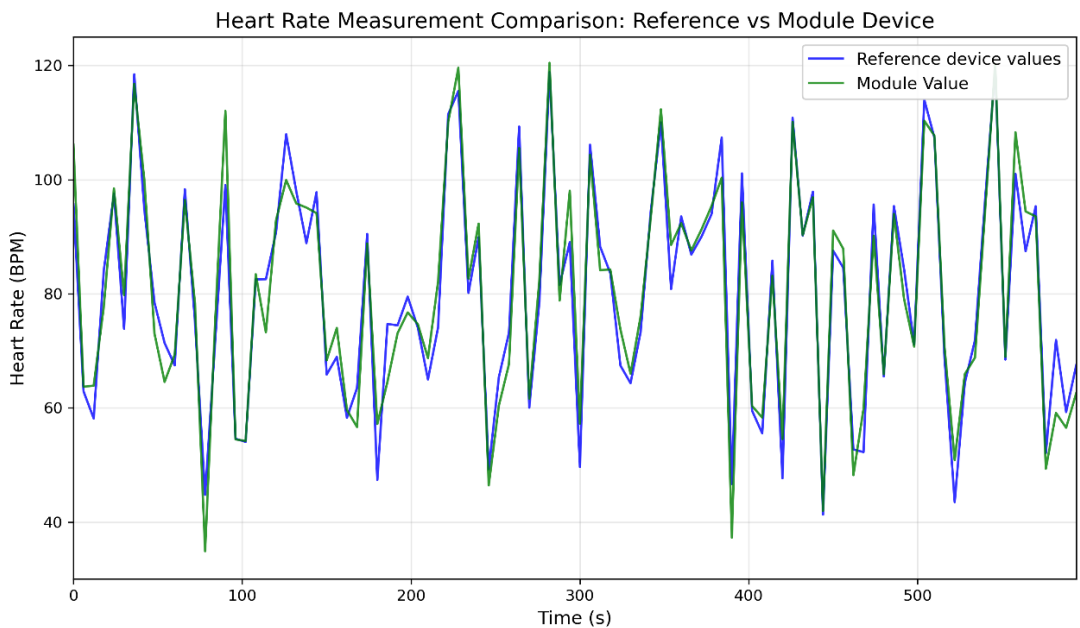

Fig. 14: Reference device values vs Module value comparison (Heart Rate BPM)

For most of the measurement period, the Heart rate fluctuated between 60-100 beats per minute. The MAX30100 showed captured BPM as low as 50-60, in the lower range and also up to 115-120, in the higher range. MAX30100 proved its effectiveness in measuring both gradual and rapid changes.

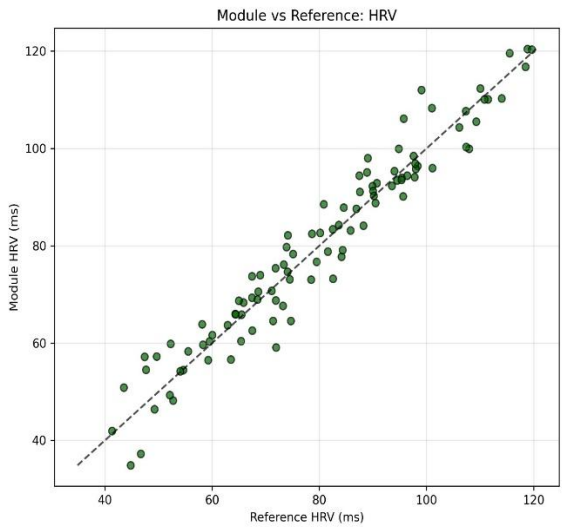
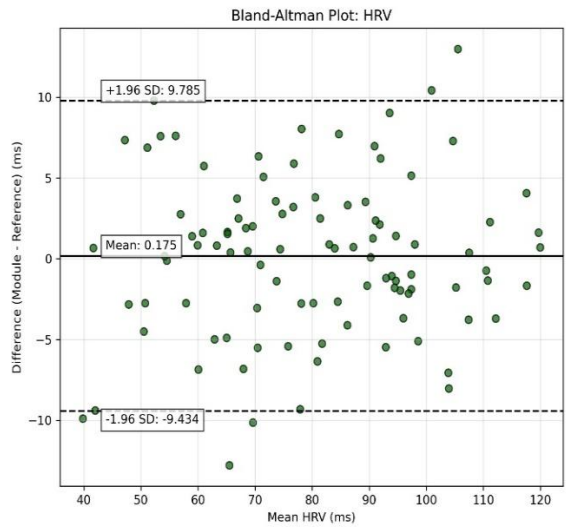

(a) Scatter plot for Heart rate BPM against medical grade devices

(b) Mean and Standard Deviation plot

Fig. 15: Bland-Altman plot for removing bias and error reduction (Heart-rate BPM)

The central line in the Bland-Altman curve implies zero difference, as it is very close. The dotted lines with standard deviation indicator show that the difference fall within acceptable clinical limits (≤2 bpm).

### 4.2.2 Body temperature and oxygen saturation

Participants in the trial were aged between 50 and 80, each age range with a normal body temperature by convention. Temperature data shown from taking 80 readings from 20 patients showed an average temperature of 98°F, with the younger demographic (50–60-year-olds) in the spectrum showed slightly lower temperature than the older counterparts (70–80-year-olds). The temperature readings were taken during specific time intervals for each patient, morning, noon, afternoon and evening. Similarly, the cohort also participated in $SpO_2$ measurement and readings were collected in same time intervals as Temperature.

To ensure that the prototype measures vitals in accordance to a clinically approved system and the values are unbiased, error/noise free, we compared the prototype values with medical grade temperature and $SpO_2$ Heart-rate BPM measurement devices. For $SpO_2$ measurement, the hospitals use JUMPER Pulse Oximeter (JPD-500D) and temperatures are captured using Genial T81 IR thermometer. Both devices are medical grade and FDA approved for hospital use.

Figures 16 and 18, respectively, shows the body temperature and oxygen saturation graph as well the real-time numeric value indicator and alert panel coupled with threshold value comparator.

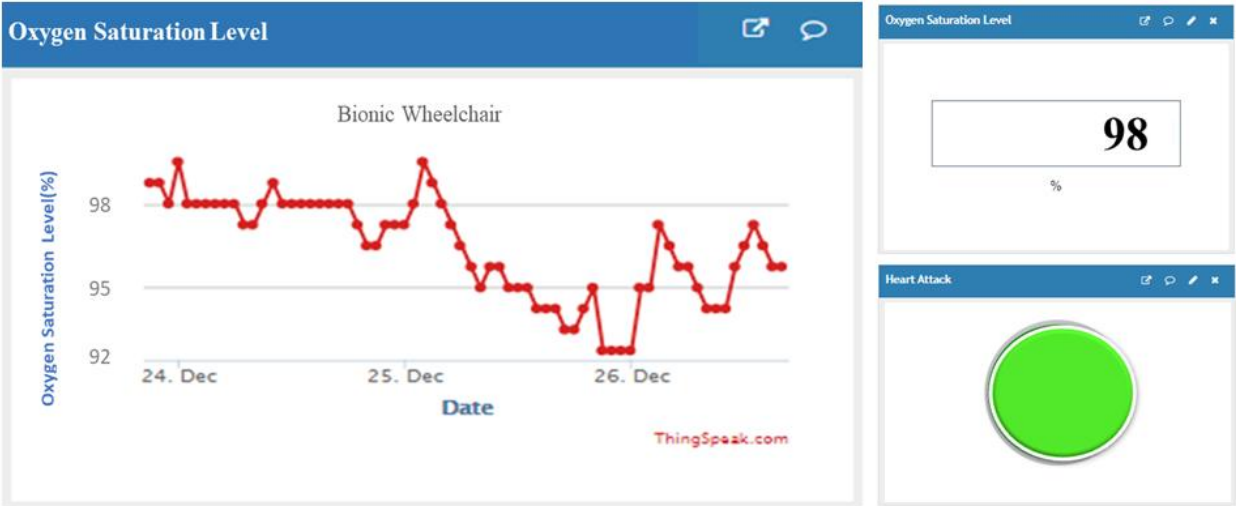

Fig. 16: Oxygen saturation level with critical condition indicator in ThingSpeak server

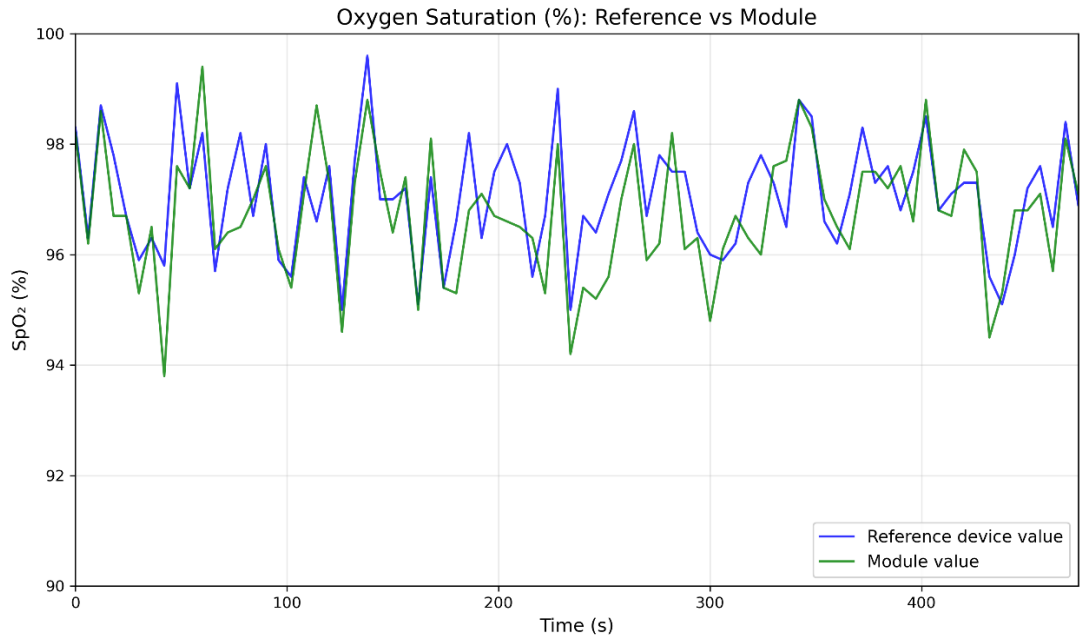

Fig. 17: Reference device vs Module values for Oxygen saturation

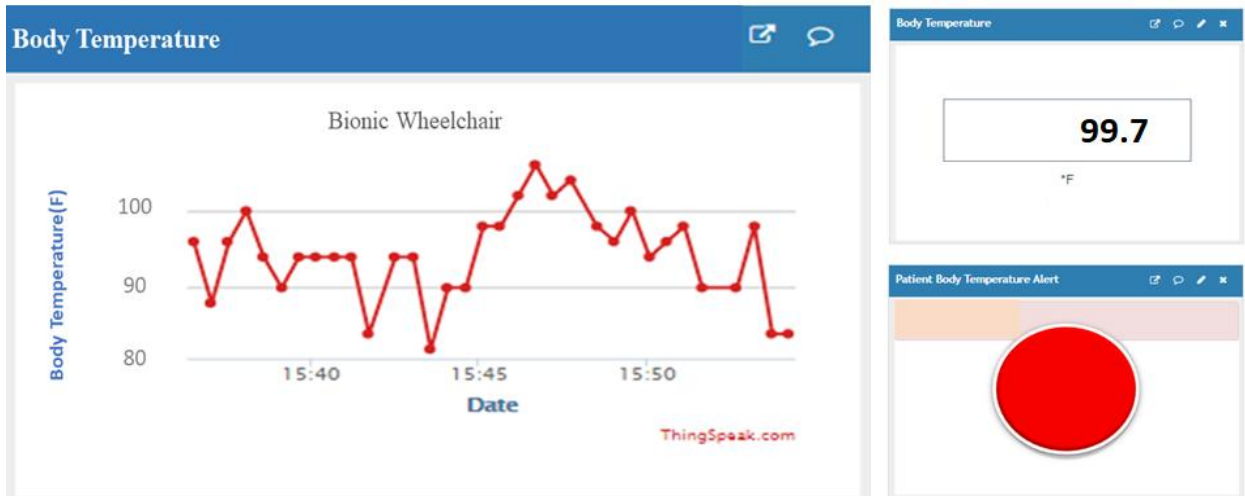

Fig. 18: Body temperature data monitoring using ThingSpeak server

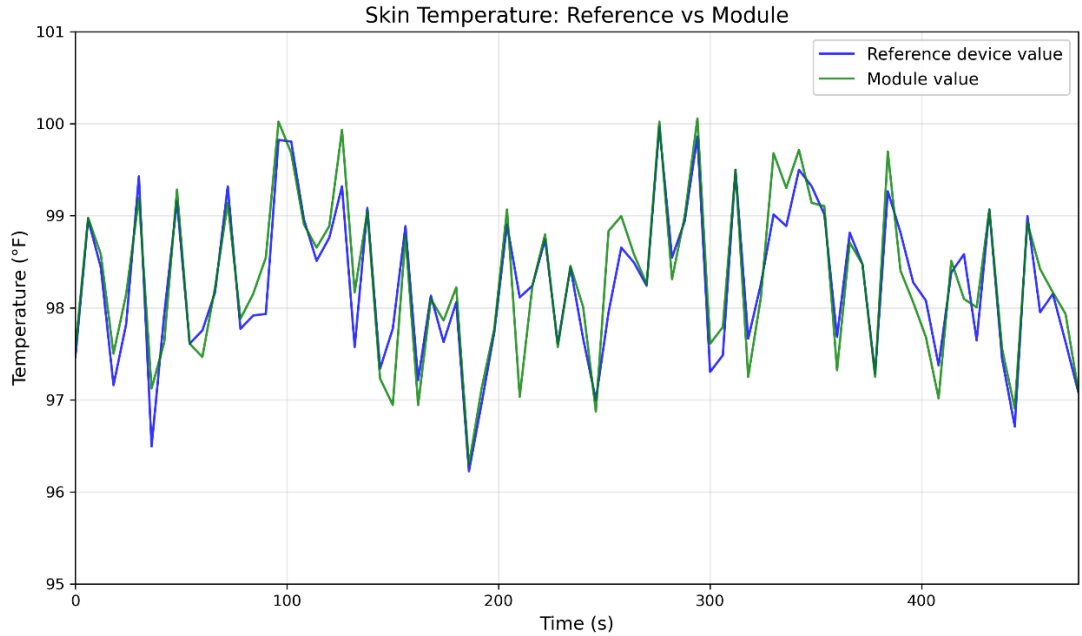

Fig. 19: Reference device values vs Module value (Temperature)

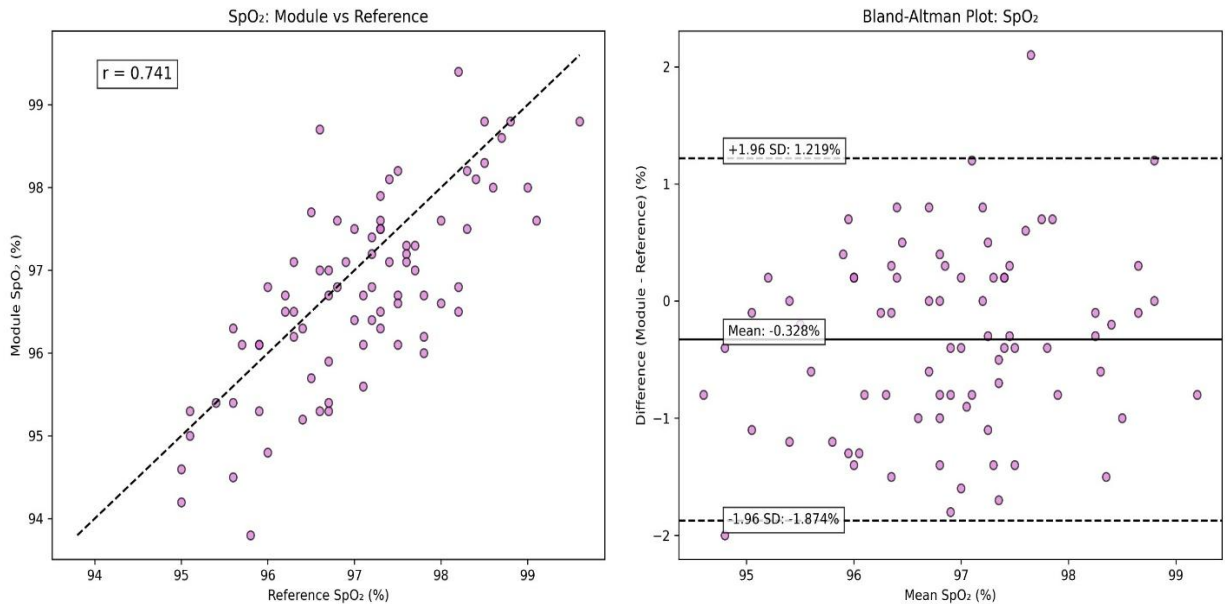

(a) Scatter plot for SpO$_2$ against medical grade devices

(b) Mean and Standard Deviation plot

Fig. 20: Bland-Altman plot for removing bias and error reduction for SpO$_2$ value

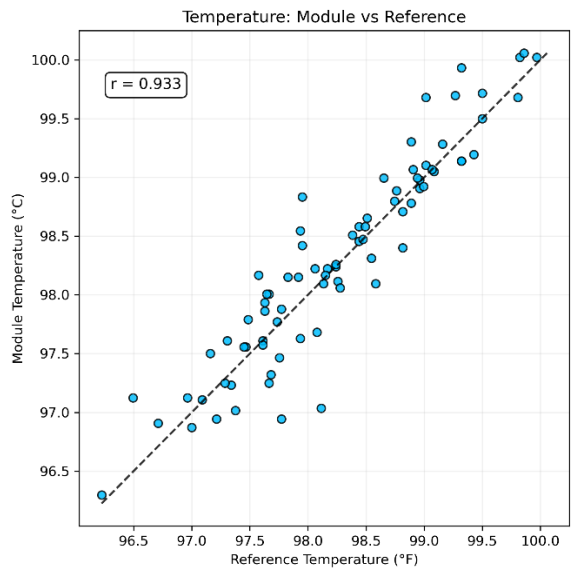 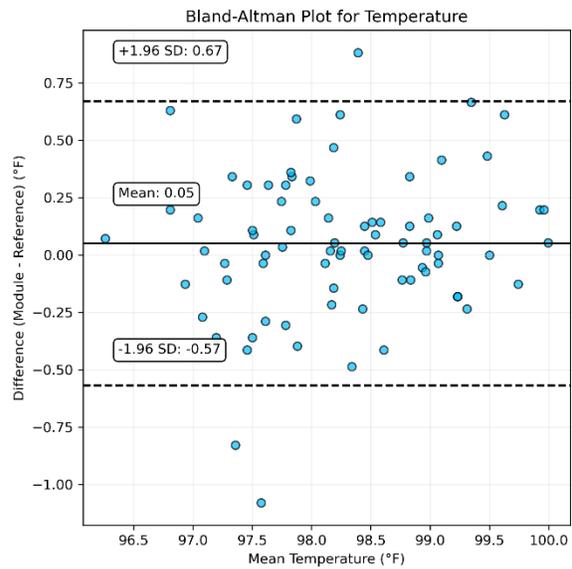

(a) Scatter plot for Temperature against medical grade devices

(b) Mean and Standard Deviation plot

Fig. 21: Bland-Altman plot to remove bias and quantify error for temperature reading

The left-hand panels plot each module reading against their corresponding reference value. The dashed line with a 45° angle is included for visual representation. The clusters around this line indicate linear relationship.

Random error and bias are quantified using Bland-Altman analysis in the right panels. Dot-wise module-reference are plotted against the pair's mean, solid central line is the mean bias and the dotted lines highlight the 95% limits of agreement.

SpO$_2$ points span across 93%-100%, with majority falling within the ±1 % of the identity. The cluster centers closely to the dashed line, showcasing a small intercept shift.

Temperature data cover 35.3–38.1 °C, falling very close to the Bland-Altman equality line. This shows the negligible systematic offset.

Bland results demonstrate the comparison with reference value readings with precision and satisfy proper clinical accuracy limits for both pulse oximetry and temperature values.

### 4.2.3 Fall detection and Convulsion

Figure 11 shows the digital value for fall detection in real time. The curve only shows falls detected by sensor and shows that value in 0 for stable or safe position and 1 for fall detected. Also, there is an email alert along with the unique condition-based alert system to instantly notify the patients' attendants of their sedentary concern.

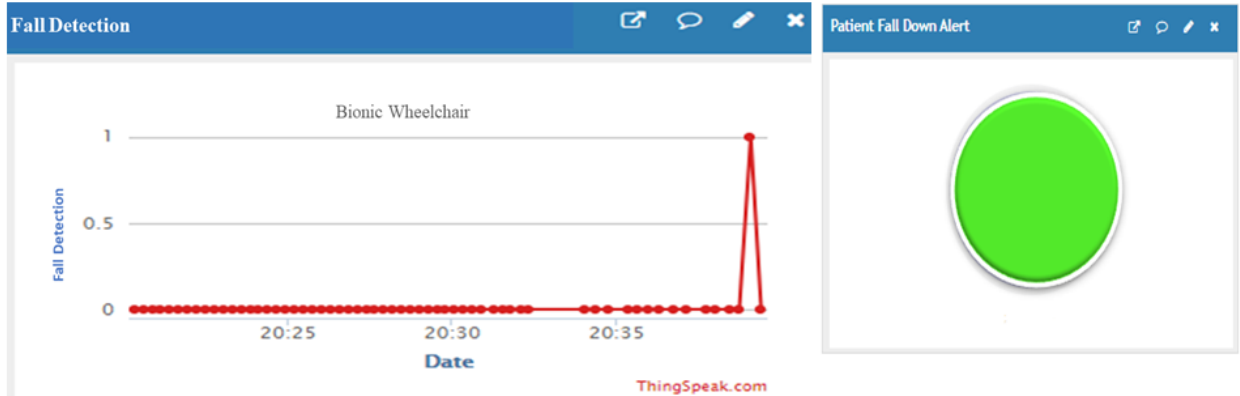

Fig. 22: Fall detection monitoring and alert system in ThingSpeak server

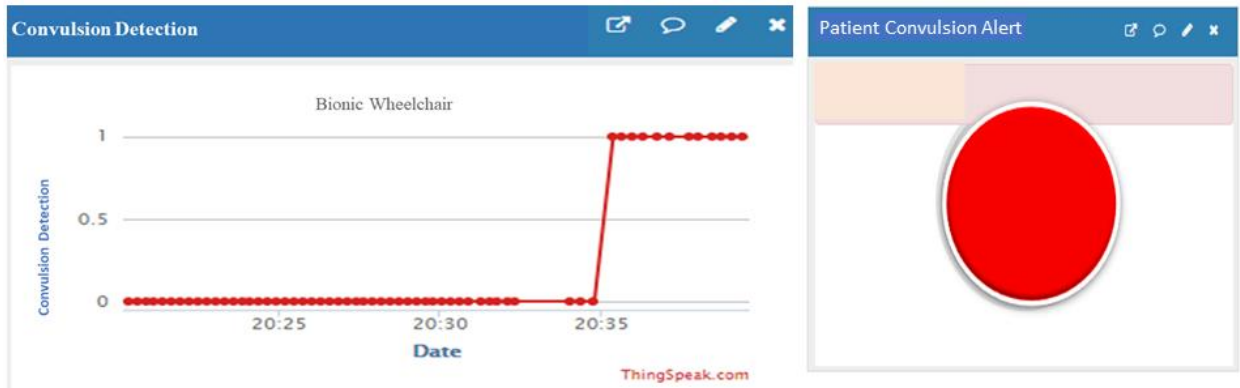

Fig. 23: Convulsion detection monitoring and alert system in ThingSpeak server

### 5. Conclusion and Future work

This study presents a robust and versatile autonomous robotic wheelchair integrated with a patient monitoring system, addressing critical challenges faced by individuals with mobility impairments. The wheelchair offers multiple control options—joystick, gesture, voice, and eye movement—making it adaptable to users with varying disabilities. Furthermore, the IoT-enabled health monitoring system ensures continuous tracking of key physiological parameters and alerts caregivers during emergencies such as falls, convulsions, or cardiac anomalies. The system's high accuracy in both mobility control and health monitoring highlights its potential to enhance user independence and safety. Despite its promising capabilities, there is scope for further improvement. Future work could focus on integrating embedded machine learning to make the system more adaptive and intelligent. For instance, predictive analytics could be used to anticipate user intentions or detect subtle health deterioration patterns. Expanding the system's usability to accommodate diverse environments and testing with a larger user base will enhance its reliability and generalizability. Additionally, incorporating energy-efficient components and exploring alternative power sources could improve the system's sustainability. With these enhancements, the proposed solution can evolve into a more comprehensive assistive technology platform, benefiting a broader range of users. The modular and IoT-enabled architecture of the system provides several advantages: Real-Time Monitoring: Continuous monitoring of health parameters ensures that caregivers are immediately informed of any critical changes. Customizability: The system's design accommodates users with different levels and types of mobility impairments. Scalability: The integration of IoT allows for easy expansion, including the addition of new sensors or features in the future. Remote Accessibility: Caregivers can monitor users from any location, reducing the need for constant physical supervision.

# References

[1] K. Zovko, L. Šerić, T. Perković, H. Belani, and P. Šolić, "IoT and health monitoring wearable devices as enabling technologies for sustainable enhancement of life quality in smart environments," *J Clean Prod*, vol. 413, p. 137506, Aug. 2023, doi: 10.1016/J.JCLEPRO.2023.137506.

[2] S. Chinnaperumal *et al.*, "Secure and intelligent 5G-enabled remote patient monitoring using ANN and Choquet integral fuzzy VIKOR," *Sci Rep*, vol. 15, no. 1, pp. 1–32, Dec. 2025, doi: 10.1038/S41598-025-93829-1;SUBJMETA=308,639,692,700,705;KWRD=HEALTH+CARE,MATHEMATICS+AND+COMPUTING,MEDICAL+RESEARCH.

[3]     I. Abdellatif, "Towards a novel approach for designing smart classrooms," *2019 IEEE 2nd International Conference on Information and Computer Technologies, ICICT 2019*, pp. 280–284, May 2019, doi: 10.1109/INFOCT.2019.8711355.

[4]     V. Bhardwaj, A. Anooja, L. S. Vermani, Sunita, and B. K. Dhaliwal, "Smart cities and the IoT: an in-depth analysis of global research trends and future directions," *Discover Internet of Things*, vol. 4, no. 1, pp. 1–21, Dec. 2024, doi: 10.1007/S43926-024-00076-3/FIGURES/2.

[5]     F. M. Talaat, R. M. El-Balka, S. Sweidan, S. A. Gamel, and A. M. Al-Zoghby, "Smart traffic management system using YOLOv11 for real-time vehicle detection and dynamic flow optimization in smart cities," *Neural Comput Appl*, pp. 1–18, Jul. 2025, doi: 10.1007/S00521-025-11434-9/METRICS.

[6]     Y. T. Ting and K. Y. Chan, "Optimising performances of LoRa based IoT enabled wireless sensor network for smart agriculture," *J Agric Food Res*, vol. 16, p. 101093, Jun. 2024, doi: 10.1016/J.JAFR.2024.101093.

[7]     S. K. Maurya, O. P. Pal, and K. Sarvakar, "Layered architecture of IoT," *Secure and Intelligent IoT-Enabled Smart Cities*, pp. 164–194, Apr. 2024, doi: 10.4018/979-8-3693-2373-1.CH009.

[8]     H. J. El-Khozondar *et al.*, "A smart energy monitoring system using ESP32 microcontroller," *e-Prime - Advances in Electrical Engineering, Electronics and Energy*, vol. 9, p. 100666, Sep. 2024, doi: 10.1016/J.PRIME.2024.100666.

[9]     A. Pagano, D. Croce, I. Tinnirello, and G. Vitale, "A Survey on LoRa for Smart Agriculture: Current Trends and Future Perspectives," *IEEE Internet Things J*, vol. 10, no. 4, pp. 3664–3679, Feb. 2023, doi: 10.1109/JIOT.2022.3230505.

[10]    S. Z. Khan, Y. Le Moullec, and M. M. Alam, "An NB-IoT-Based Edge-of-Things Framework for Energy-Efficient Image Transfer," *Sensors (Basel)*, vol. 21, no. 17, p. 5929, Sep. 2021, doi: 10.3390/S21175929.

[11]    N. Verma, S. Singh, and D. Prasad, "A Review on existing IoT Architecture and Communication Protocols used in Healthcare Monitoring System," *Journal of The Institution of Engineers (India): Series B*, vol. 103, no. 1, pp. 245–257, Feb. 2022, doi: 10.1007/S40031-021-00632-3/METRICS.

[12]    S. Messinis, N. Temenos, N. E. Protonotarios, I. Rallis, D. Kalogeras, and N. Doulamis, "Enhancing Internet of Medical Things security with artificial intelligence: A comprehensive review," *Comput Biol Med*, vol. 170, p. 108036, Mar. 2024, doi: 10.1016/J.COMPBIOMED.2024.108036.

[13] I. Singh, S. Kumar, and J. Koh, "Innovations in wearable biosensors: A pathway to 24/7 personalized healthcare," *Measurement*, vol. 254, p. 117938, Oct. 2025, doi: 10.1016/J.MEASUREMENT.2025.117938.

[14] S. N. A. B. M. Nashruddin, F. H. M. Salleh, R. M. Yunus, and H. B. Zaman, "Artificial intelligence−powered electrochemical sensor: Recent advances, challenges, and prospects," *Heliyon*, vol. 10, no. 18, p. e37964, Sep. 2024, doi: 10.1016/J.HELIYON.2024.E37964.

[15] J. Mao *et al.*, "A Health Monitoring System Based on Flexible Triboelectric Sensors for Intelligence Medical Internet of Things and its Applications in Virtual Reality," Sep. 2023, Accessed: Aug. 03, 2025. [Online]. Available: https://arxiv.org/pdf/2309.07185

[16] "Ageing and health." Accessed: Aug. 03, 2025. [Online]. Available: https://www.who.int/news-room/fact-sheets/detail/ageing-and-health

[17] F. Béthoux, P. Calmels, and V. Gautheron, "Changes in the quality of life of hemiplegic stroke patients with time: A preliminary report," *Am J Phys Med Rehabil*, vol. 78, no. 1, pp. 19–23, Jan. 1999, doi: 10.1097/00002060-199901000-00006,.

[18] R. A. Robbins, Z. Simmons, B. A. Bremer, S. M. Walsh, and S. Fischer, "Quality of life in ALS is maintained as physical function declines," *Neurology*, vol. 56, no. 4, pp. 442–444, Feb. 2001, doi: 10.1212/WNL.56.4.442,.

[19] F. Möller, R. Rupp, N. Weidner, C. Gutenbrunner, Y. B. Kalke, and R. F. Abel, "Long term outcome of functional independence and quality of life after traumatic SCI in Germany," *Spinal Cord*, vol. 59, no. 8, pp. 902–909, Aug. 2021, doi: 10.1038/S41393-021-00659-9;SUBJMETA=1824,375,578,617,692,699;KWRD=SPINAL+CORD+DISEASES,TRAUMA.

[20] "Almost one billion children and adults with disabilities and older persons in need of assistive technology denied access, according to new report." Accessed: Aug. 03, 2025. [Online]. Available: https://www.who.int/news/item/16-05-2022-almost-one-billion-children-and-adults-with-disabilities-and-older-persons-in-need-of-assistive-technology-denied-access--according-to-new-report

[21] B. Woods and N. Watson, "A short history of powered wheelchairs," *Assistive Technology*, vol. 15, no. 2, pp. 164–180, 2003, doi: 10.1080/10400435.2003.10131900;PAGE:STRING:ARTICLE/CHAPTER.

[22] H. Wang, B. Salatin, G. G. Grindle, D. Ding, and R. A. Cooper, "Real-time model based electrical powered wheelchair control," *Med Eng Phys*, vol. 31, no. 10, pp. 1244–1254, Dec. 2009, doi: 10.1016/J.MEDENGPHY.2009.08.002.

[23] S. P. Levine, D. A. Bell, and Y. Koren, "NavChair: An example of a shared-control system for assistive technologies," pp. 136–143, 1994, doi: 10.1007/3-540-58476-5_116.

[24] I. Ulrich and I. Nourbakhsh, "Appearance-based place recognition for topological localization," *Proc IEEE Int Conf Robot Autom*, vol. 2, pp. 1023–1029, 2000, doi: 10.1109/ROBOT.2000.844734.

[25] Z. Wang, J. Xu, J. Zhang, R. Slavin, and D. Zhu, "An intelligent assistive driving solution based on smartphone for power wheelchair mobility," *Journal of Systems Architecture*, vol. 149, p. 103105, Apr. 2024, doi: 10.1016/J.SYSARC.2024.103105.

[26] A. Palumbo *et al.*, "An Innovative Device Based on Human-Machine Interface (HMI) for Powered Wheelchair Control for Neurodegenerative Disease: A Proof-of-Concept," *Sensors 2024, Vol. 24, Page 4774*, vol. 24, no. 15, p. 4774, Jul. 2024, doi: 10.3390/S24154774.

[27] S. A. H. Hussain, I. Raza, S. A. Hussain, M. H. Jamal, T. Gulrez, and A. Zia, "A mental state aware brain computer interface for adaptive control of electric powered wheelchair," *Sci Rep*, vol. 15, no. 1, Dec. 2025, doi: 10.1038/S41598-024-82252-7,.

[28] S. A. H. Hussain, I. Raza, S. A. Hussain, M. H. Jamal, T. Gulrez, and A. Zia, "A mental state aware brain computer interface for adaptive control of electric powered wheelchair," *Sci Rep*, vol. 15, no. 1, Dec. 2025, doi: 10.1038/S41598-024-82252-7,.

[29] "Smart Wheelchair Market Size & Growth Forecast 2024-2034." Accessed: Aug. 03, 2025. [Online]. Available: https://www.futuremarketinsights.com/reports/smart-wheelchair-market

[30] J. Xu, Z. Huang, L. Liu, X. Li, and K. Wei, "Eye-Gaze Controlled Wheelchair Based on Deep Learning," *Sensors 2023, Vol. 23, Page 6239*, vol. 23, no. 13, p. 6239, Jul. 2023, doi: 10.3390/S23136239.

[31] R. Meligy, A. R. Ahmad, and S. Mekid, "An IoT-Based Smart Wheelchair with EEG Control and Vital Sign Monitoring," *Engineering Proceedings 2024, Vol. 82, Page 46*, vol. 82, no. 1, p. 46, Nov. 2024, doi: 10.3390/ECSA-11-20489.

[32] S. A. H. Hussain, I. Raza, S. A. Hussain, M. H. Jamal, T. Gulrez, and A. Zia, "A mental state aware brain computer interface for adaptive control of electric powered wheelchair," *Sci Rep*, vol. 15, no. 1, pp. 1–14, Dec. 2025, doi: 10.1038/S41598-024-82252-7;SUBJMETA=308,639,692,700,705;KWRD=HEALTH+CARE,MATHEMATICS+AND+COMPUTING,MEDICAL+RESEARCH.

[33] V. Ramaraj, A. Paralikar, E. J. Lee, S. M. Anwar, and R. Monfaredi, "Development of a Modular Real-time Shared-control System for a Smart Wheelchair," *J Signal Process Syst*, vol. 96, no. 3, pp. 203–214, Mar. 2024, doi: 10.1007/S11265-022-01828-6/METRICS.

[34] L. Hou, J. Latif, P. Mehryar, A. Zulfiqur, S. Withers, and A. Plastropoulos, "IoT Based Smart Wheelchair for Elderly Healthcare Monitoring," *2021 IEEE 6th International Conference on Computer and Communication Systems, ICCCS 2021*, pp. 917–921, Apr. 2021, doi: 10.1109/ICCCS52626.2021.9449273.

[35] M. A. Hossain, M. F. K. Khondakar, M. H. Sarowar, and M. J. U. Qureshi, "Design and Implementation of an Autonomous Wheelchair," *2019 4th International Conference on Electrical Information and Communication Technology, EICT 2019*, Dec. 2019, doi: 10.1109/EICT48899.2019.9068851.

**Declaration of generative AI and AI-assisted technologies in the writing process**

During the preparation of this work the author(s) used ChatGPT in order to improve readability and language of manuscript. After using this tool/service, the author(s) reviewed and edited the content as needed and take(s) full responsibility for the content of the publication